\documentclass[sort&compress, numafflabel]{elsarticle}

\usepackage[]{natbib}

\usepackage{enumitem}
\usepackage[breaklinks,hidelinks]{hyperref}
\usepackage{times}
\usepackage{latexsym}

\usepackage[margin=1in]{geometry}
\usepackage{float}
\usepackage{subfiles}
\usepackage{multirow}
\usepackage{array}
\usepackage{hyphenat}
\usepackage{subcaption}
\usepackage{longtable}
\usepackage{graphicx}
\usepackage{adjustbox}
\usepackage{enumitem}
\usepackage{url}
\usepackage{csquotes}

\usepackage[color=yellow, textsize=small]{todonotes}

\usepackage{bm}
\usepackage{booktabs}
\usepackage{float}
\usepackage[english]{babel}
\usepackage{blindtext}
\usepackage{textcomp}
\usepackage{soul,color}

\makeatletter
\def\ps@pprintTitle{%
 \let\@oddhead\@empty
 \let\@evenhead\@empty
 \def\@oddfoot{}%
 \let\@evenfoot\@oddfoot}
\makeatother

\usepackage{titlesec}
\titleformat{\section}
      {\normalfont\bfseries}
      {\thesection}
      {0ex}
      {\MakeUppercase}

\titleformat{\subsection}
      {\normalfont\bfseries}
      {\thesection}
      {0ex}
      {}

\titleformat{\subsubsection}
      {\normalfont}
      {\thesection}
      {0ex}
      {}

\usepackage{setspace}
\newif\ifsubfile
\subfiletrue

\newif\iftif
\tiffalse
\usepackage{microtype}

\usepackage{bm}

\usepackage{array}
\usepackage{amssymb}
\usepackage{pifont}
\newcommand\blfootnote[1]{%
  \begingroup
  \renewcommand\thefootnote{}\footnote{#1}%
  \addtocounter{footnote}{-1}%
  \endgroup
}

\title{CACER: Clinical Concept Annotations for Cancer Events and Relations}

\author[add1]{Yujuan Velvin Fu\textsuperscript{*}\blfootnote{\textsuperscript{*}Corresponding author: Yujuan Velvin Fu, velvinfu@uw.edu.}}
\ead{}
\author[add2]{Giridhar Kaushik Ramachandran}
\ead{}
\author[add3]{Ahmad Halwani}
\ead{}
\author[add4]{Bridget T. McInnes}
\ead{}
\author[add5]{Fei Xia}
\ead{}
\author[add2]{Kevin Lybarger}
\ead{}
\author[add1]{Meliha Yetisgen}
\ead{melihay@uw.edu}
\author[add2]{Özlem Uzuner}
\ead{}

\address[add1]{Department of Biomedical Informatics \& Medical Education, University of Washington, Seattle, WA, USA}
\address[add2]{Department of Information Sciences and Technology, George Mason University, Fairfax, VA, USA} 
\address[add3]{Huntsman   Cancer   Institute, University of Utah,  Salt Lake City, UT,  USA }
\address[add4]{Department of Computer Science, Virginia Commonwealth University, Richmond, VA, USA}
\address[add5]{Department of Linguistics, University of Washington, Seattle, WA, USA}

\begin{document}

\subfilefalse

\newpageafter{author}

\begin{abstract}
\noindent\textbf{Objective:}
Clinical notes contain unstructured representations of patient histories, including the relationships between medical problems and prescription drugs. To investigate the relationship between cancer drugs and their associated symptom burden, we extract structured, semantic representations of medical problem and drug information from the clinical narratives of oncology notes.\\

\noindent\textbf{Materials and Methods:} 
We present Clinical Concept Annotations for Cancer Events and Relations (CACER), a novel corpus with fine-grained annotations for over 48,000 medical problems and drug events and 10,000 drug-problem and problem-problem relations. Leveraging CACER, we develop and evaluate transformer-based information extraction (IE) models such as BERT, Flan-T5, Llama3, and GPT-4 using fine-tuning and in-context learning (ICL). \\

\noindent\textbf{Results:} 
In event extraction, the fine-tuned BERT and Llama3 models achieved the highest performance at 88.2-88.0 F1, which is comparable to the inter-annotator agreement (IAA) of 88.4 F1. In relation extraction, the fine-tuned BERT, Flan-T5, and Llama3 achieved the highest performance at 61.8-65.3 F1. GPT-4 with ICL achieved the worst performance across both tasks. \\

\noindent\textbf{Discussion:} 
The fine-tuned models significantly outperformed GPT-4 in ICL, highlighting the importance of annotated training data and model optimization. Furthermore, the BERT models performed similarly to Llama3. For our task, LLMs offer no performance advantage over the smaller BERT models.\\

\noindent\textbf{Conclusions:}
We introduce CACER, a novel corpus with fine-grained annotations for medical problems, drugs, and their relationships in clinical narratives of oncology notes. State-of-the-art transformer models achieved performance comparable to IAA for several extraction tasks. 
\\
\end{abstract}



\maketitle

\pagebreak
\section*{Background and Significance}
Clinical notes capture detailed descriptions of patient status and disease progression from the care provider's perspective through unstructured text \citep{ledade2017narrative}. In oncology, these narratives are comprehensive and cover diverse symptoms \citep{deshields2014persistence}, multiple drug cycles, and side effects \citep{schirrmacher2019chemotherapy}. Such unstructured notes contain valuable information that complements structured data in electronic health records (EHRs) \citep{ehrenstein2019obtaining}, such as symptoms \citep{zhou2023generalizing}, diagnostic hypotheses, and treatment decisions. Understanding these clinical narratives is crucial for treatment decisions \citep{jensen2017analysis}, patient management, and quality assurance.

Natural language processing (NLP) methods for information extraction (IE) can convert unstructured narratives into structured data \citep{wang2018clinical}, enabling large-scale, real-time use of those rich information in clinical decision support applications and generation of real-world evidence in learning health systems. High-performing IE models require advanced techniques and annotated datasets for training and evaluation.
Existing research has produced systems that extract information about cancer diagnoses and treatments \citep{datta2019frame}; however, a significant gap remains. Specifically, existing frameworks do not fully capture the relationships between cancer diagnoses, symptoms, and medications. This gap highlights the need for a unified approach for characterizing medical problems and drug information, including their interconnections.

Our annotation focuses on two cancer populations: prostate cancer and diffuse large B-Cell lymphoma (DLBCL).
Prostate cancer is the second most common cancer among men and exhibits considerable heterogeneity \cite{siegel2022cancer,rawla2019epidemiology}. Some patients have a less aggressive, chronic form, whereas others face a highly aggressive disease linked to increased morbidity and mortality, necessitating more intensive treatments \cite{rawla2019epidemiology}. Similarly, DLBCL represents the most common aggressive lymphoma \cite{susanibar20212021}. Together, these cancers exemplify the spectrum of chronic and aggressive cancer trajectories.

\subsection*{Medical Problem and Drug Data Sets}
Clinical IE includes a variety of classification tasks, including text classification, relation extraction (RE), and event extraction (EE) \cite{wang2018clinical}. We utilize EE to characterize medical problems and drugs, and RE to determine the relationships between them. Each event includes a trigger representing a clinical concept (problem or drug) and fine-grained attributes (e.g., assertion or anatomy) \citep{datta2019frame}. Previous research on event-based medical problem extraction either (1) focuses on a subset of problems, such as symptoms \citep{zhou2023generalizing} or diseases and disorders \citep{pradhan2014semeval}, or (2) lacks granular annotation. For instance, the 2010 i2b2/VA challenge \citep{uzuner20112010} only included assertion attributes for medical problems (present vs. absent), without detailing severity or anatomy. Cancer-focused EE studies \citep{zeng2017contralateral, coden2009automatically, breischneider2017automatic} often overlook key factors, like symptoms (excluding pain) \cite{heintzelman2013longitudinal} and comorbidities \cite{ping2013information}, which are essential for understanding diagnosis and treatment. We present a comprehensive annotation schema that captures all medical problems \citep{lybarger2021extracting, turner2022domain, zhou2023generalizing}.

The clinical relationships between drugs and medical problems inform treatment and diagnosis, but are often complex. Most existing literature narrowly focuses on a subset of possible relations, including adverse drug events \cite{henry20202018, jagannatha2019overview} and clinical temporal relationships \cite{sun2013evaluating, viani2019annotating, bethard2016semeval} in clinical notes and gene-cancer interactions in biomedical literature \citep{kawashima2017text, cao2022novel}. The 2010 i2b2/VA challenge annotated six detailed relations between medical problems and treatments in discharge summaries, which we use to characterize interactions between medical problems and drugs in clinical narratives of oncology notes.

\subsection*{Clinical IE Approaches}
IE allows for the secondary use of clinical narratives in clinical and translational research \citep{wang2018clinical}, as well as near real-time EHR clinical decision support functionalities. It can enhance understanding of drug discontinuation, symptom monitoring, and adverse event management \citep{alkaitis2021automated, dimartino2022identification, lindvall2022deep, nishioka2023adverse}. Most clinical IE approaches employ separate models for event and relationship extraction in multi-step processes. These models have progressed from rule-based systems \cite{aronson2001effective, savova2010mayo, soysal2018clamp} and feature-engineered models \cite{culotta2004dependency, culotta-etal-2006-integrating, sahu2016relation} to neural networks, culminating in transformer architectures \citep{vaswani2017attention, landolsi2023information}. Bidirectional Encoder Representations from Transformers (BERT) \citep{devlin2018bert}, an encoder-only model, has shown superior performance in clinical IE \cite{landolsi2023information}. Clinical variants of BERT \citep{alsentzer2019publicly} have achieved high performance in clinical IE tasks, like drug-problem RE \citep{roy2021incorporating}. Some BERT-based architectures, like Packed Levitated Marker (PL-Marker) \cite{ye2022plmarker}, adopt a two-step approach for extracting spans (arguments) and relationships (argument roles), including IE for medical imaging reports \cite{park2024novel}. BERT architectures with multiple output layers, like Span-based Event and Relation Transformer (SpERT) \citep{eberts2019SpERT}, can jointly extract spans (arguments) and relationships (argument roles), including the extraction of social determinants of health \citep{Lybarger2022mspERT}.

Recent progress in generative language models (GLMs) includes encoder-decoder models, like the Fine-tuned Language Net Text-To-Text Transfer Transformer (Flan-T5) \citep{chung2022scaling}, and decoder-only models, like the Large Language Model Meta AI (LLaMA) \citep{touvron2023llama} and Generative Pre-trained Transformers 4 (GPT-4) \citep{openai2023gpt4}. GLMs have set new performance standards in various clinical benchmark tasks. Smaller models like T5 \cite{2020t5} have been fine-tuned for specialized tasks such as medical document classification, named entity recognition (NER) \citep{lu2022clinicalt5}, and medical database query generation \citep{dobbins2023leafai}. GLMs with billions or trillions of parameters, such as GPT-4, are called Large Language Models (LLMs) and excel at in-context learning (ICL) \cite{zhao2023survey}. ICL allows these models to adapt to tasks based solely on prompts without changing their parameter weights and has been successfully applied to clinical IE tasks \citep{ramachandran-etal-2023-prompt, fu2024extracting, hu2024zero}.

\section*{Objective}
We introduce \textbf{C}linical Concept \textbf{A}nnotations for \textbf{C}ancer \textbf{E}vents and \textbf{R}elations (CACER), a novel corpus of cancer patient clinical oncology notes from Fred Hutch Cancer Center, with detailed annotations for medical problems, drugs, and their relationships. We benchmark this dataset with high-performing models and outline future directions. Key contributions include:
\begin{itemize}
    \item We provide comprehensive, fine-grained annotations for 48k medical problem and drug events and 10k drug-problem and problem-problem relations. CACER will be made available to the research community pending institutional review board approval.
    
    \item We develop state-of-the-art extractors using BERT models and GLMs via supervised fine-tuning and ICL approaches. For GLMs, we explore various prompting strategies, including Question-Answering (QA). RE encompasses long-distance, inter-sentence relationships.

\end{itemize}

\section*{Materials and Methods}
\subsection*{Creating the CACER Corpus}
\subsubsection*{Data Set Collection}
We used a clinical data set of outpatient records for two types of cancer, prostate cancer and DLBCL, from the Fred Hutch Cancer Center from 2015-2018. 
This data set includes 1,453 prostate cancer patients (with 10k notes) and 818 patients with DLBCL (with 11k notes). CACER consists of 575 clinical notes randomly sampled from this data set. Following de-identification, we randomly sampled notes containing over 30 lines and manually excluded clinical notes that were duplicates, EHR templates, or overlapped with other notes. We divided the data into training (400 notes), validation (60 notes), and test (115 notes) subsets, ensuring that no patient was included in multiple subsets. The Supplementary Files provides detailed dataset statistics and patient demographics.

\subsubsection*{Event Annotation Schema}
The CACER annotation schema encompasses medical problem (\textit{Problem}) and drug (\textit{Drug}) events, along with the relations between them. It builds upon the schema used in our previous research on COVID-19 symptoms \citep{lybarger2021extracting} and lung and ovarian cancer symptoms \citep{turner2022domain}. We expanded the symptom annotation guidelines to encompass all medical problems and drug events in the cancer domain. 



\begin{figure}[h]
\centering
\includegraphics[width=9cm]{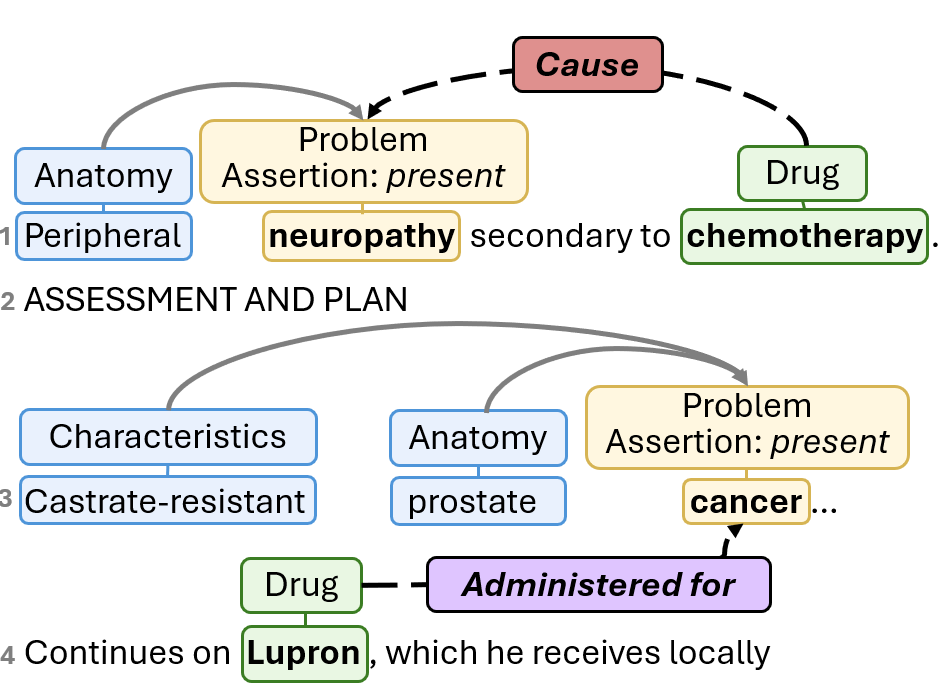}
\caption{ Annotation example from CACER. line 1 includes the intra-sentence relation, \textit{Causes} and line 2-4 contains the inter-sentence relation, \textit{administered for}. The inter-sentence relation is indicated by the note section subtitle, `ASSESSMENT AND PLAN', linking the treatment to the main diagnosis.}\label{fig:brat}
\end{figure}

Each \textit{Problem} event is marked by a \textit{trigger} and multiple attributes (i.e., \textit{arguments}). The \textit{Problem} trigger is a text span that most clearly and concisely expresses the medical problem. 
An argument has a name (i.e., `argument type") such as \textit{Duration} and a value that corresponds to a text span such as `three months".  For some argument types such as \textit{Assertion}, their values can be normalized into a pre-defined set of subtypes (e.g., `present", `absent").
For instance, the subtype `present" might correspond to text spans such as `is observed" or `was found", 
or it is simply implied by the occurrence of a disease name. 
We call the argument with a subtype label a `\textit{labeled}" argument and the one without a `\textit{span-only}" argument. Figure \ref{fig:brat} provides examples for both types of arguments.
\textit{Drug} events only include a trigger, without any arguments. We will address the potential for incorporating more fine-grained drug annotations within oncology notes in the discussion section.
Table \ref{tab:event_sheme} summarizes the event annotation schema.  Our annotation guidelines will be made available on the project's GitHub page.

\begin{table*}[]
\centering
\begin{tabular}{cccrrrcc}
\hline
\multirow{2}{*}{\textbf{Event}} & \multirow{2}{*}{\textbf{\begin{tabular}[c]{@{}c@{}}Trigger\\ \& Arg.\end{tabular}}} & \multirow{2}{*}{\textbf{\begin{tabular}[c]{@{}c@{}}Trigger examples\\ \& Argument subtypes\end{tabular}}} & \multicolumn{3}{c}{\textbf{\# Labels}} & \multicolumn{2}{c}{\textbf{IAA F1}} \\ \cline{4-8} 
 &  &  & \textbf{Train} & \textbf{Valid} & \textbf{Test} & \textbf{\begin{tabular}[c]{@{}c@{}}Without\\ preAnn\end{tabular}} & \textbf{\begin{tabular}[c]{@{}c@{}}With\\ preAnn\end{tabular}} \\ \hline
Drug & Trigger* & `ibuprofen', `lupron', ... & 11,118 & 2,104 & 3,534 & 97.5 & 97.1 \\ \cline{1-3}
\multirow{8}{*}{Problem} & Trigger* & `cancer', `vomitting'... & 21,453 & 3,575 & 6,555 & 89.6 & 96.5 \\ \cline{2-3}
 & Assertion* & \begin{tabular}[c]{@{}c@{}}\{present, hypothetical, \\ absent, conditional, \\ possible, not\_patient\}\end{tabular} & 21,453 & 3,575 & 6,555 & 87.2 & 94.0 \\ \cline{2-3}
 & Change & \begin{tabular}[c]{@{}c@{}}\{worsening, no\_change,\\ improving, resolved\}\end{tabular} & 1,440 & 293 & 418 & 78.4 & 85.1 \\ \cline{2-3}
 & Severity & \{mild, moderate, severe\} & 775 & 168 & 254 & 84.6 & 89.8 \\ \cline{2-3}
 & Anatomy & `prostate', `back', ... & 9,880 & 1,638 & 2,877 & 82.9 & 91.9 \\ \cline{2-3}
 & Characteristics & `recurrant', `metastatic', ... & 4,749 & 830 & 1,439 & 66.2 & 87.2 \\ \cline{2-3}
 & Duration & `1 year', `two weeks', ... & 930 & 171 & 298 & 92.9 & 81.7 \\ \cline{2-3}
 & Frequency & `every day', `rarely', ... & 245 & 35 & 79 & 92.3 & 69.6 \\ \cline{1-3}
Overall & - & - & 72,043 & 12,389 & 22,009 & 88.4 & 94.1 \\ \hline
\end{tabular}
\caption{Event schema, statistics, and inter-annotator agreement (IAA). * indicates required trigger and argument annotations for each event. Argument subtypes are enclosed within \{ \}. CACER contains a total of 48,339 events, which is equivalent to  the number of triggers. The columns \textit{Without preAnn} and \textit{With preAnn} show the IAA without pre-annotations (49 notes) and with pre-annotations (113 notes).}
\label{tab:event_sheme}
\end{table*}

\subsubsection*{Relation Annotation Schema}
We use six \textit{Drug}-\textit{Problem} and \textit{Problem}-\textit{Problem} relation types from the 2010 i2b2/VA corpus \cite{uzuner20112010}, as shown in Table \ref{tab:relation_schema}. Each relation includes a head, a tail, and a relation type. For \textit{Drug}-\textit{Problem} relations, the head is a \textit{Drug} trigger, and the tail is a \textit{Problem} trigger. For \textit{Problem}-\textit{Problem} relations, both the head and the tail are \textit{Problem} triggers, where the tail \textit{Problem} describes characteristics of the head \textit{Problem} or identifies it as the cause or superclass of the tail.

\begin{table*}[h]
\centering
\begin{tabular}{llrrrcrrrc}
\hline
\multirow{3}{*}{\textbf{Relation}} & \multirow{3}{*}{\textbf{Head event}} & \multicolumn{4}{c}{\textbf{All relations}} & \multicolumn{4}{c}{\textbf{Intra-sentence relations}} \\ \cline{3-10} 
 &  & \multicolumn{3}{c}{\textbf{\# labels}} & \multirow{2}{*}{\textbf{\begin{tabular}[c]{@{}c@{}}IAA\\ F1\end{tabular}}} & \multicolumn{3}{c}{\textbf{\# labels}} & \multirow{2}{*}{\textbf{\begin{tabular}[c]{@{}c@{}}IAA\\ F1\end{tabular}}} \\ \cline{3-5} \cline{7-9}
 &  & \textbf{Train} & \textbf{Valid} & \textbf{Test} &  & \textbf{Train} & \textbf{Valid} & \textbf{Test} &  \\ \hline
AdminFor & \multirow{5}{*}{Drug} & 3,715 & 762 & 1,422 & 75.4 & 2,456 & 373 & 783 & 86.9 \\
NotAdminBeause &  & 130 & 49 & 54 & 50.6 & 106 & 37 & 44 & 50.0 \\
Causes &  & 729 & 232 & 321 & 75.1 & 644 & 199 & 278 & 75.7 \\
Improves &  & 502 & 87 & 133 & 62.5 & 342 & 53 & 65 & 71.1 \\
Worsens &  & 257 & 68 & 121 & 43.0 & 199 & 52 & 70 & 47.2 \\ \cline{1-2}
\begin{tabular}[c]{@{}l@{}}Problem-Indicates\\-Problem (PIP)\end{tabular} & Medical problem & 1,257 & 259 & 350 & 54.1 & 1,141 & 238 & 304 & 54.5 \\ \cline{1-2}
Overall & - & 6,590 & 1,457 & 2,402 & 69.6 & 4,888 & 952 & 1,544 & 74.6 \\ \hline
\end{tabular}

\caption{Relation schema, statistics, and IAA. All relation tails are medical problems. CACER contains a total of 10,449 relations. The IAA is derived from the 162 doubly annotated notes.}
\label{tab:relation_schema}
\end{table*}
Relation annotation requires either: (1) explicit linguistic cues indicating a relationship or (2) implicit relationships based on medical knowledge. Most relations are intra-sentence and are based on verb phrases like `prescribed for.'  Inter-sentence relations can be inferred by section headers, e.g., `Cancer Treatment -\textgreater ... on \textless DATE\textgreater, Lupron is given." indicates an \textit{AdminFor} relation between cancer and lupron. Otherwise, they are frequently implicit and rely on general medical knowledge. A given \textit{Problem} or \textit{Drug} may be mentioned multiple times in a note, but only the closest pairs of triggers are annotated. We define the relation \textit{context window} as the smallest continuous sequence of sentences that includes both head and tail triggers. In CACER, $<1\%$ of the relations have a context window longer than five sentences.

Annotation is carried out using a web-based annotation tool, BRAT \citep{stenetorp2012brat}. Example annotations are shown in Figure \ref{fig:brat}. 

\subsection*{Building IE Systems}
The CACER IE task comprises multiple subtasks. The extraction of \textit{Drug} and \textit{Problem} events requires the identification of triggers and argument spans, prediction of the relationships (argument roles) between triggers and arguments, and resolution of the subtype labels for labeled arguments. Additionally, the relations between \textit{Drug} and \textit{Problem} must be predicted.  The subtasks can be performed through multi-step extraction approaches, as well as end-to-end approaches that jointly extract all phenomena. Our experiments used BERT models and GLMs, incorporating both fine-tuning and ICL. Hyperparameters, such as batch size, gradient accumulation steps, and number of epochs, were optimized on the validation set, and the best-performing models were then evaluated on the test set. Model performance was assessed using a one-sided, bootstrap T-test with 10,000 iterations, where a sample is a note, the p-value threshold is 0.05. 

\begin{figure}[h]
\centering
\includegraphics[width=14cm]{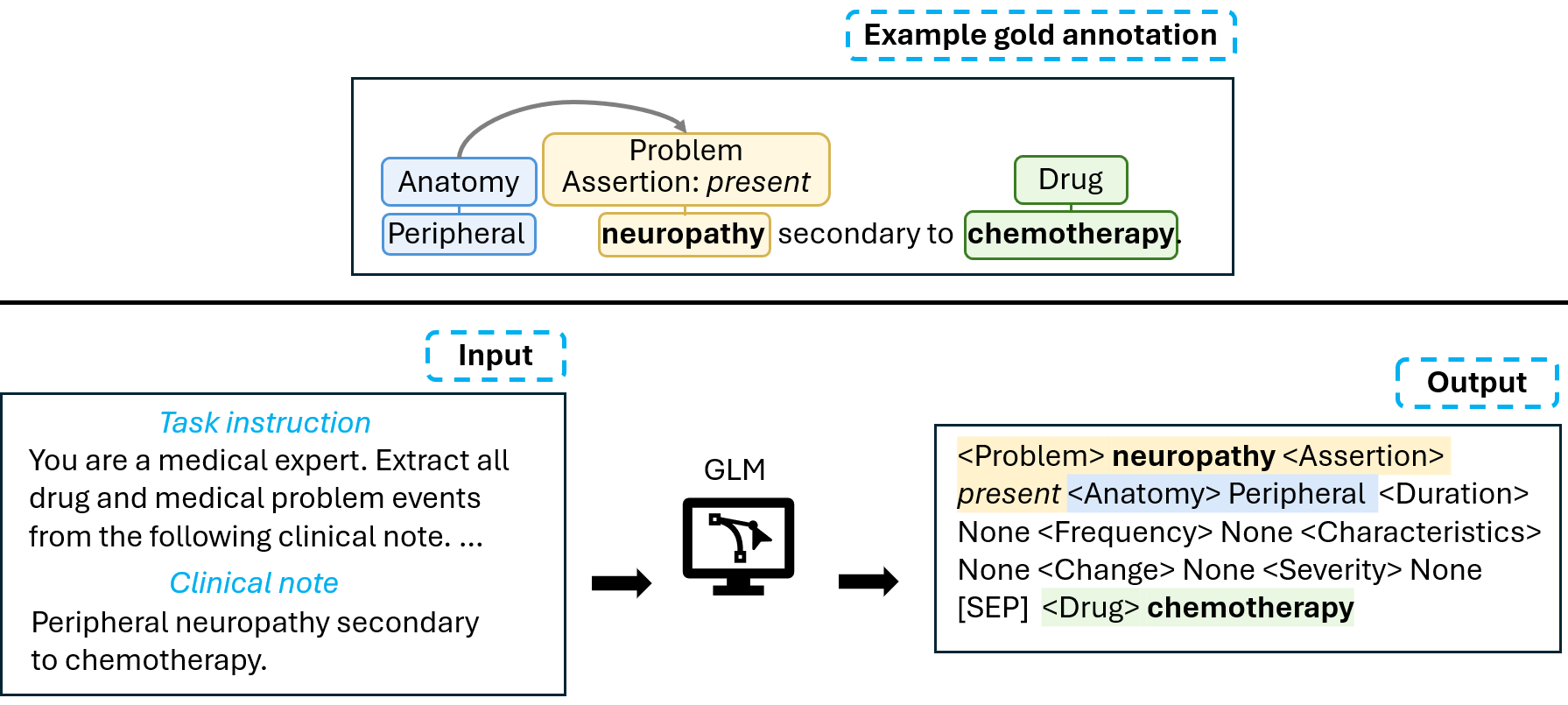}
\caption{Input and output formats for GLMs in EE.}\label{fig:nlg_EE}
\end{figure}

For BERT-based encoder models, we selected SpERT \citep{eberts2019SpERT} and PL-Marker \citep{ye2022plmarker}, both using Bio+ClinicalBERT \citep{alsentzer-etal-2019-publicly}, pre-trained on biomedical and clinical texts, as their encoder. For GLMs, we fine-tuned Flan-T5-large \citep{chung2022scaling} and Llama3-8B-instruct \citep{llama3} and used GPT-4 in an ICL setting \citep{openai2023gpt4}. All GLMs used a consistent prompt format, as shown in Figure \ref{fig:nlg_EE} and Figure \ref{fig:nlg_RE}. For fine-tuned GLMs, we applied parameter-efficient fine-tuning (PEFT) with low-rank adaptation (LoRA) \cite{hu2021lora}. GPT-4 experiments were conducted in a Health Insurance Portability and Accountability Act-compliant Azure environment.

\subsubsection*{Event Extraction (EE)}

Almost all event arguments co-occur in the same sentence as the trigger, so EE models operate on sentences. The GLM event extraction used a common format \citep{romanowski2023extracting, fu2024extracting}, where the input prompt contains a task description and the target sentence, and the output is a structured text representation of the extracted events ordered by the location of the trigger. If no events are extracted, the output is `None'. Each event is presented in a uniform template comprising a trigger and all arguments. Individual triggers or arguments start with a type label, followed by a subtype label or spans. Multiple spans of the same argument type are separated by the token \textless  s\textgreater, as in `\textless  Problem\textgreater\  pain \textless  Assertion\textgreater\  present \textless  Anatomy\textgreater\  back \textless  s\textgreater\  neck \textless  Duration\textgreater\ ...'. Events are separated by the [SEP] token. To be considered valid, all predicted text spans must occur exactly as they do in the original sentence. GPT-4 experimentation utilized ICL, where the prompt included a concise summary of the annotation guidelines and two randomly selected in-context sentence examples (one containing both \textit{Problem} and \textit{Drug} events and another without events). The complete EE prompts and condensed annotation guideline can be found in the Supplementary File. 

\subsubsection*{Relation Extraction (RE)}
\begin{figure}[h]
\centering
\includegraphics[width=14cm]{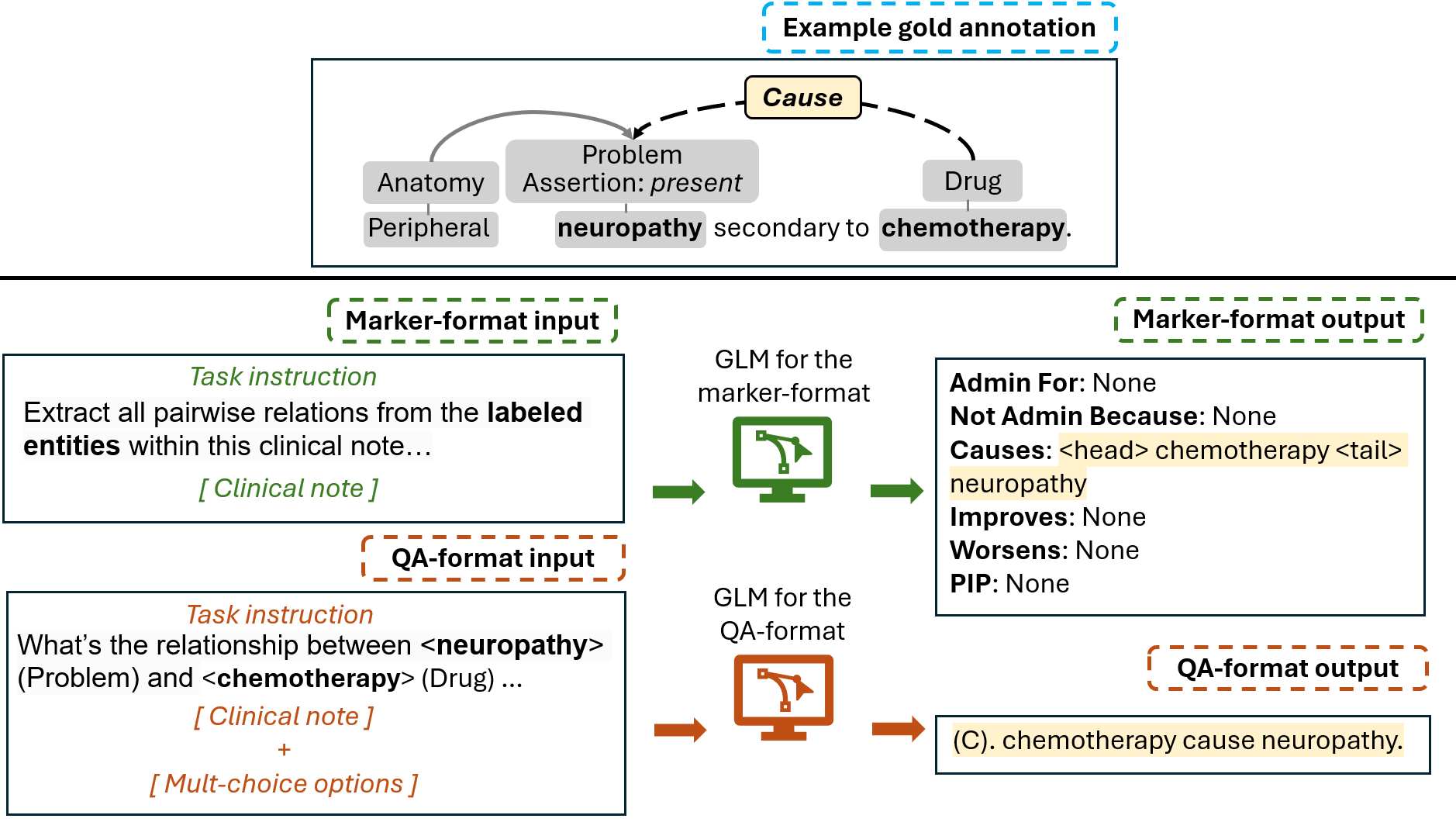}
\caption{Input and output formats for GLMs in RE.}\label{fig:nlg_RE}
\end{figure}

\label{section:RE_method}
RE is unconstrained by sentence boundaries, and relations may span the entire note. We define the context window for a relation as the smallest continuous sequence of sentences that includes both head and tail events. We consider the RE input to be all context windows that are smaller than 400 BERT tokens and five sentences. This approach covers 98.7\% of all relations. Details of the context window implementation are in the Supplementary File.

Figure \ref{fig:nlg_RE} presents the two RE formats for GLMs: \textit{Marker} and \textit{QA}. The Marker format aligns with PL-Marker, using special markers to denote \textit{Drug} and \textit{Problem} triggers in the input. The model outputs one relation type per line, returning a default \textit{None} label if no relation is identified. If multiple instances of a relation type are found, they are separated by a [SEP] token. To avoid duplication, only relations extracted from their specific context windows are considered valid predictions, even when there is overlap between intra- and inter-sentence relations. The QA format identifies relations through multiple-choice questions, following previous work \citep{singhal2022large}. For each possible event pair in a context window, a unique input prompt is created that includes task instructions, the context window with labeled event pairs, and the potential relations based on condensed annotation guidelines. There are five possible \textit{Drug}-\textit{Problem} relations, one asymmetric \textit{Problem}-\textit{Problem} relation, and the option to indicate no relation by `None of the above'. The complete RE prompts can be found in the Supplementary File. 

\subsection*{Inter-annotator agreement (IAA) and IE Evaluation} \label{evaluation_criteria}
We calculated IAA and evaluated the performance of our IE systems using precision, recall, and F1. For events, we employed a relaxed evaluation similar to the N2C2 SDoH challenge \citep{lybarger20232022}. Two triggers are \textit{equivalent} if they have the same event type and overlapping spans. Two labeled arguments are equivalent if (1) their argument types match, (2) their subtype labels match, and (3) they are attached to equivalent triggers\footnote{For labeled arguments, the subtype label normalizes its corresponding span into a categorized value. Therefore, we focus on the subtype label and do not additionally evaluate the argument span.}. Span-only arguments are considered equivalent if (1) their argument types match, (2) their spans overlap, and (3) they are attached to equivalent triggers. For relations, equivalence is established if (1) the head and tail triggers are equivalent and (2) the relation types match.

\section*{Results}

\subsection*{CACER Annotation}

CACER was annotated by three University of Washington (UW) medical students. To familiarize themselves with the schema and refine the guidelines, annotators independently annotated five notes per training session, across a total of six such sessions. Across the annotation rounds,  events were annotated first, followed by annotation of relations between events. IAA was calculated per round to monitor quality, and disagreements were discussed. The training set was singly annotated, 78\% of the validation set was doubly annotated, and the test set was fully doubly annotated.

In the first two rounds, annotators demonstrated consistent annotation of triggers, enabling the development of high-performing pre-annotation models. 
Previous research has shown that using pre-annotation can speed up the gold standard development for clinical NER, while maintaining similar IAA \cite{lingren2014evaluating}.
We trained a pre-annotation model on 29 notes from rounds 1-2 to assist annotators in focusing on more complex tasks and accelerate annotation. The pre-annotation model was a multi-label variation of SpERT (mSpERT) \citep{eberts2019SpERT, Lybarger2022mspERT}, built with Bio+Clinical BERT \cite{alsentzer-etal-2019-publicly}. We evaluated the pre-annotation model on four singly-annotated notes containing 293 \textit{Problem} and 174 \textit{Drug} triggers, achieving 85.2 and 83.3 F1, respectively. Starting in round 3, notes were pre-annotated with triggers and arguments, but without the trigger-argument linkages. Annotators reviewed and corrected these pre-annotations, and added the trigger-argument linkages in the process.

Table \ref{tab:event_sheme} presents the IAA for events with and without the use of pre-annotation. Pre-annotation improved the overall IAA and the IAA for \textit{Problem} triggers but did not impact the IAA for \textit{Drug} triggers or some \textit{Problem} arguments. Annotators reported pre-annotation was helpful as it saved time, which is consistent with the findings from previous research \cite{lingren2014evaluating}. However, we observed a decline in IAA for the two less frequent \textit{Problem} arguments, \textit{Duration} and \textit{Frequency}.  This decline is likely due to the limited training data (29 notes) used to create the pre-annotation classifier and the linguistic variability in the expression of these terms, such as `at night' for \textit{Frequency}. The pre-annotation classifier has a higher rate of false negatives for these two arguments, resulting in some missed annotations and lower IAA. To address these issues and improve IAA, we reviewed disagreements during feedback discussions after each annotation round. Additionally, all test samples were doubly annotated to ensure annotation consistency.

Table \ref{tab:relation_schema} presents the IAA for relations, focusing on intra-sentence relations, where sentence boundaries are defined using SpaCy (en\_core\_web\_sm) \footnote{https://spacy.io/}. Intra-sentence relations account for 70.7\% of relations. The micro-average relation IAA is 74.6 F1 for intra-sentence relations and 69.6 F1 for all relations. Inter-sentence relations pose significant challenges due to their implicit expression, making them more difficult to identify and define. To enhance the RE IAA, we conducted a quality check on singly annotated notes using an RE model to identify annotation inconsistencies. We trained the FLAN-T5 model using the GLM-QA format with gold standard events, as described in the following Section,  Relation Extraction, on singly annotated notes from one annotator and then used to predict results for notes annotated by another. A third annotator reviewed and corrected any discrepancies between the model and human labels.

\subsection*{EE Results}
\begin{table}[]
\centering
\begin{tabular}{lcrccccc}
\hline
\multirow{2}{*}{\textbf{Event}} & \multirow{2}{*}{\textbf{\begin{tabular}[c]{@{}c@{}}Trigger\\ \& Arg.\end{tabular}}} & \multirow{2}{*}{\textbf{\begin{tabular}[c]{@{}c@{}}\#\\ Labels\end{tabular}}} & \multicolumn{2}{c}{\textbf{BERT-based LM}} & \multicolumn{3}{c}{\textbf{GLM}} \\ \cline{4-8} 
 &  &  & \textbf{SpERT} & \textbf{PL-Marker} & \textbf{Flan-T5} & \textbf{\begin{tabular}[c]{@{}c@{}}Llama 3\\ 8B\end{tabular}} & \textbf{\begin{tabular}[c]{@{}c@{}}GPT-4\\ (ICL)\end{tabular}} \\ \hline
Drug & Trigger & 3,534 & \underline{94.0} & \multicolumn{1}{c|}{\underline{93.9}} & \underline{91.4} & \underline{\textbf{95.0}} & 82.9 \\
Problem & Trigger & 6,555 & \underline{93.1} & \multicolumn{1}{c|}{\underline{93.1}} & \underline{90.9} & \underline{\textbf{93.2}} & 68.2 \\
Problem & Assertion & 6,555 & \underline{89.3} & \multicolumn{1}{c|}{\underline{89.6}} & 86.4 & \underline{\textbf{89.8}} & 62.7 \\
Problem & Change & 418 & \underline{72.0} & \multicolumn{1}{c|}{\underline{71.5}} & \underline{\textbf{75.2}} & \underline{73.0} & 35.0 \\
Problem & Severity & 254 & \underline{\textbf{74.7}} & \multicolumn{1}{c|}{\underline{73.3}} & \underline{69.7} & \underline{70.5} & 35.7 \\
Problem & Anatomy & 2,877 & \underline{81.7} & \multicolumn{1}{c|}{\underline{\textbf{83.0}}} & 67.4 & \underline{79.9} & 57.0 \\
Problem & Characteristics & 1,439 & \underline{71.9} & \multicolumn{1}{c|}{\underline{\textbf{75.0}}} & 57.5 & \underline{68.8} & 17.0 \\
Problem & Duration & 298 & \underline{\textbf{72.9}} & \multicolumn{1}{c|}{\underline{72.7}} & \underline{67.5} & \underline{69.6} & 22.4 \\
Problem & Frequency & 79 & 64.6 & \multicolumn{1}{c|}{62.2} & 65.3 & \textbf{65.6} & 45.1 \\ \hline
Overall & -- & 22,009 & \underline{88.3} & \multicolumn{1}{c|}{\underline{\textbf{88.8}}} & 83.9 & \underline{88.2} & 61.7 \\ \hline
\end{tabular}
\caption{Event extraction F1 performance. Except for GPT-4 (ICL), all approaches are finetuned on the CACER train set. 
\textbf{Bold} numbers represent the highest numerical scores. \underline{Underlined} numbers denote the top-performing systems, indicating statistical significance over non-underlined systems. There is no significant difference between any of the underlined systems.
}
\label{tab:event_performance}
\end{table}
Table \ref{tab:event_performance} presents the EE performance. BERT and Llama3 show no significant difference in overall F1 scores but significantly outperform Flan-T5 and GPT-4 (ICL). The top three models achieve performance close to IAA (without pre-annotation) for triggers, frequent arguments, and overall performance. However, for less frequent arguments (\textit{Severity}, \textit{Duration}, and \textit{Frequency}), the best models underperform compared to IAA. For GLMs, EE performance decreases in sentences with multiple events due to the generation of longer texts, increasing the likelihood of cascading errors from previously generated tokens. We noted that GPT-4 (ICL) significantly underperforms other models in almost every category of trigger and argument performance, except for the infrequent argument, \textit{Frequency}. This underperformance is primarily due to GPT-4 (ICL)'s occasional non-compliance with in-context annotation guideline instructions. For instance, GPT-4 (ICL) tends to merge attributes such as \textit{Characteristics} and \textit{Anatomy} into \textit{Problem} triggers, resulting in the omission of meaningful arguments.

\subsection*{EE Error Analysis}
\begin{table}[h]
\centering
\resizebox{\columnwidth}{!}{%
\begin{tabular}{llccccc}
\hline
\multirow{2}{*}{\textbf{\begin{tabular}[c]{@{}l@{}}Error\\ Types\end{tabular}}} & \multirow{2}{*}{\textbf{\begin{tabular}[c]{@{}l@{}}Error\\ Subtypes\end{tabular}}} & \multicolumn{5}{c}{\textbf{Language Models}} \\ \cline{3-7} 
 &  & \textbf{SpERT} & \textbf{PL-Marker} & \textbf{Flan-T5} & \textbf{Llama 3 8B} & \textbf{GPT-4 (ICL)} \\ \hline
\multirow{6}{*}{\textbf{\begin{tabular}[c]{@{}l@{}}False\\ Negatives\end{tabular}}} & Abnormal measures & 12 & 3 & 2 & 4 & 0 \\
 & Uppercases/Abbreviations & 5 & 4 & 2 & 4 & 2 \\
 & Headers & 18 & 24 & 16 & 14 & 6 \\
 & Long spans & 6 & 6 & 4 & 0 & 0 \\
 & Dense events & 0 & 0 & 42 & 14 & 0 \\
 & Others & 24 & 28 & 42 & 32 & 78 \\ \cline{1-2}
\multirow{3}{*}{\textbf{\begin{tabular}[c]{@{}l@{}}False \\ Positives\end{tabular}}} & Overlapping spans & 25 & 0 & 0 & 0 & 0 \\
 & Hallucinations & 0 & 0 & 0 & 0 & 23 \\
 & Others & 32 & 25 & 26 & 18 & 101 \\ \cline{1-2}
\multirow{3}{*}{\textbf{\begin{tabular}[c]{@{}l@{}}Mis-classi-\\ fication\end{tabular}}} & Arguments & 4 & 2 & 2 & 2 & 12 \\
 & Labels & 6 & 4 & 4 & 2 & 10 \\
 & Switched events and arguments & 2 & 6 & 4 & 5 & 12 \\ \cline{1-2}
\textbf{Overall} & - & 134 & 102 & 144 & 95 & 244 \\ \hline
\end{tabular}
}
\caption{Types of RE errors in 5 sampled notes from the test set. The 5 notes have 987 event triggers and arguments in total. The category, dense events, refers to the scenario where locations of multiple events are in close proximity. The category, \textit{switched events and arguments}, refers to the scenario where an event trigger is classified as an argument, or vice versa.}
\label{tab:ee_error}
\end{table}

We conducted an error analysis by randomly sampling 5 notes from the test set and manually characterizing the associated EE errors. Table \ref{tab:ee_error} presents a distribution of error types by LM. 

There are diverse sources of false negatives (FNs), including: (1) abnormal test results (FN \textit{Problem} events) such as `\textit{PSA started rising up to 9.2}'; (2) words in all caps, to express emphasis or represent abbreviations and acronyms such as `\textit{ATIVAN}', `\textit{VALIUM}', `\textit{SVI}, `\textit{ECE}' (casing issues may be mitigated with case-normalization in preprocessing); (3) \textit{Anatomy} arguments in section headers such as '\textit{Genitourinary}:  no rash/erythema in groin area'; (4) long \textit{Problem} triggers, such as `\textit{activity is pretty much confined to walking to the bathroom from the bed}'. (5) locations of multiple events in close proximity (dense events), which are usually from a procedural checklist such as `he denied any significant symptoms including \textit{headache}, \textit{loss of consciousness}, \textit{vision change}, \textit{chest pain}, ...'. There can be more than 15 such \textit{Problem} events in a single sentence, and Flan-T5 and Llama 3 8B can fail to capture the last few events.

As a frequent source of false positives (FPs), SpERT tends to classify multiple overlapping spans. For example, in the phrase `\textit{red raised rash}', SpERT classifies three \textit{Characteristics} arguments, `\textit{red}', `\textit{raised}', and `\textit{red raised}'. These overlap errors could be reduced through post-processing by merging the overlapping spans of the same type. GPT-4 tends to generate hallucinations for labeled arguments, which are either nonsensical or not accurately reflective of the source content provided \cite{ji2023survey}. For example, GPT-4 generates the \textit{Severity} argument label, \textit{severe}, for the event `\textit{metastasis cancer}', even though no descriptions of its severity are involved. The other FPs are frequently associated with nuanced discrepancies to the annotation task definition, and we find many common FPs among multiple LMs. For example, the clinical measure, `\textit{PSA nadir}' is misclassified as a \textit{Problem} event; `happened last night' is a single time point but misclassified as a \textit{Duration} argument. Those types of FPs are especially frequent for GPT-4. 

The mis-classifications involve (1) incorrect argument types, for example, `pain \textit{8/10}' is a \textit{Severity} argument but is predicted as \textit{Characteristics}; (2) incorrect subtype labels for labeled arguments, for example, the \textit{Assertion} in `prescribed for future \textit{pain}' should be \textit{hypothetical} but is predicted as \textit{possible}; (3) misclassification of event triggers as arguments or vice versa, for example, `bony \textit{metastasis} with concerning \textit{lesions}' should have two \textit{Problem} events (`\textit{metastasis}' and `\textit{lesions}') but `\textit{lesions}' labeled as a \textit{Characteristics} argument. 

\subsection*{RE Results}

\begin{table}[h]
\centering
\resizebox{\columnwidth}{!}{%

\begin{tabular}{lrcccccccc}
\multicolumn{10}{l}{\textbf{Table 5A,  all relations}} \\ \hline
\multirow{2}{*}{\textbf{Relation}} & \multicolumn{1}{l}{\multirow{2}{*}{\textbf{\begin{tabular}[c]{@{}l@{}}\#\\ Labels\end{tabular}}}} & \multicolumn{2}{c}{\textbf{BERT-based  LM}} & \multicolumn{3}{c}{\textbf{GLM - marker format}} & \multicolumn{3}{c}{\textbf{GLM - QA format}} \\ \cline{3-10} 
 & \multicolumn{1}{l}{} & \textbf{SpERT} & \textbf{PL-Marker} & \textbf{Flan-T5} & \textbf{\begin{tabular}[c]{@{}c@{}}Llama 3\\ 8B\end{tabular}} & \textbf{\begin{tabular}[c]{@{}c@{}}GPT-4 \\ (ICL)\end{tabular}} & \textbf{Flan-T5} & \textbf{\begin{tabular}[c]{@{}c@{}}Llama 3\\ 8B\end{tabular}} & \textbf{\begin{tabular}[c]{@{}c@{}}GPT-4 \\ (ICL)\end{tabular}} \\ \hline
AdminFor & 1,422 & - & \underline{71.9} & 56.4 & 57.4 & - & \underline{75.0} & \underline{\textbf{76.4}} & - \\
NotAdminBecause & 54 & - & 52.9 & 42.4 & 35.0 & - & \textbf{57.4} & \textbf{57.4} & - \\
Causes & 321 & - & 75.7 & 51.9 & 57.3 & - & 78.4 & \textbf{78.7} & - \\
Improves & 133 & - & \underline{56.4} & 43.2 & 43.8 & - & \underline{\textbf{63.5}} & \underline{54.3} & - \\
Worsens & 121 & - & 28.6 & 31.6 & 28.4 & - & \textbf{42.6} & 34.6 & - \\
PIP & 350 & - & \underline{\textbf{58.0}} & 38.0 & 40.1 & - & \underline{52.9} & \underline{54.3} & - \\
Overall & 2,402 & - & \underline{67.4} & 50.3 & 52.1 & - & \underline{69.2} & \underline{\textbf{70.3}} & - \\ \hline
\begin{tabular}[c]{@{}l@{}}Overall\\ (predicted events)\end{tabular} & 2,402 & \underline{61.8} & \underline{62.0} & 48.6 & 51.6 & - & \underline{62.2} & \underline{\textbf{65.3}} & \underline{-} \\ \hline
 & \multicolumn{1}{l}{} & \multicolumn{1}{l}{} & \multicolumn{1}{l}{} & \multicolumn{1}{l}{} & \multicolumn{1}{l}{} & \multicolumn{1}{l}{} & \multicolumn{1}{l}{} & \multicolumn{1}{l}{} & \multicolumn{1}{l}{} \\
\multicolumn{10}{l}{\textbf{Table 5B, intra-sentence relations}} \\ \hline
\multirow{2}{*}{\textbf{Relation}} & \multicolumn{1}{l}{\multirow{2}{*}{\textbf{\begin{tabular}[c]{@{}l@{}}\#\\ Labels\end{tabular}}}} & \multicolumn{2}{c}{\textbf{BERT-based  LM}} & \multicolumn{3}{c}{\textbf{GLM - marker format}} & \multicolumn{3}{c}{\textbf{GLM - QA format}} \\ \cline{3-10} 
 & \multicolumn{1}{l}{} & \textbf{SpERT} & \textbf{PL-Marker} & \textbf{Flan-T5} & \textbf{\begin{tabular}[c]{@{}c@{}}Llama 3\\ 8B\end{tabular}} & \textbf{\begin{tabular}[c]{@{}c@{}}GPT-4 \\ (ICL)\end{tabular}} & \textbf{Flan-T5} & \textbf{\begin{tabular}[c]{@{}c@{}}Llama 3\\ 8B\end{tabular}} & \textbf{\begin{tabular}[c]{@{}c@{}}GPT-4 \\ (ICL)\end{tabular}} \\ \hline
AdminFor & 765 & - & \underline{84.6} & 78.4 & \underline{79.8} & 71.8 & \underline{86.8} & \underline{\textbf{88.8}} & \underline{83.7} \\
NotAdminBecause & 44 & - & 55.8 & 57.1 & 50.7 & 35.4 & 61.9 & \textbf{62.5} & 36.4 \\
Causes & 274 & - & \underline{80.8} & 71.0 & \underline{75.5} & 63.1 & \underline{83.5} & \underline{\textbf{84.4}} & 74.4 \\
Improves & 65 & - & \textbf{73.8} & 65.6 & 62.1 & 45.9 & 72.5 & 68.4 & 56.7 \\
Worsens & 70 & - & 40.4 & 41.1 & 40.8 & 32.7 & \textbf{52.8} & 45.6 & 49.5 \\
PIP & 304 & - & \underline{\textbf{61.5}} & 53.8 & \underline{58.1} & 22.4 & \underline{56.4} & \underline{58.6} & 46.3 \\ \hline
Overall & 1,522 & - & \underline{76.5} & \underline{69.3} & \underline{71.7} & 56.7 & \underline{76.7} & \underline{\textbf{78.9}} & 65.9 \\ \hline
\begin{tabular}[c]{@{}l@{}}Overall\\ (predicted events)\end{tabular} & 1,522 & 70.2 & 70.6 & 64.8 & 69.4 & - & 67.9 & \textbf{73.1} & \underline{-} \\ \hline
\end{tabular}

}
\caption{Relation extraction performance. Except for GPT-4 (ICL), all approaches are finetuned on the CACER train set. Except for the two overall F1 with predicted events, all RE F1 are based on gold events. \textbf{Bold} numbers represent the highest numerical scores. \underline{Underlined} numbers denote the top-performing systems, indicating statistical significance over non-underlined systems. There was no significant difference among any of the underlined systems.}
\label{tab:rel_performance}
\end{table}

CACER includes both intra- and inter-sentence relations. Table \ref{tab:rel_performance} presents the RE performance: Table \ref{tab:rel_performance}A includes all relations, and Table \ref{tab:rel_performance}B includes only intra-sentence relations. Both tables detail overall performance assuming gold standard events and end-to-end performance using predicted events. 
In the RE tasks, Llama with a QA format achieved the highest F1 but was not significantly better than SpERT or PL-Marker. In Table \ref{tab:rel_performance}A, the QA prompting strategy surpassed the Marker approach, suggesting that the more constrained QA format is a better approach for GLMs. Additionally, breaking down complex RE tasks into individual entity-pair tasks achieves higher performance.

In Table \ref{tab:rel_performance}B, GPT-4 (ICL) exhibited significantly poorer performance than all other methods, with many false positives. A major source of false positives for \textit{Problem}-\textit{Problem} relations is the listing of multiple \textit{Problem}s, such as symptoms, adverse events, and comorbidities, in close proximity within oncology notes, without contextual descriptors or medical knowledge to indicate actual relationships. GPT-4 inference incurs higher computational and financial costs, so we did not evaluate GPT-4 (ICL) further in other experiment settings. 

For intra-sentence RE with gold standard events in Table \ref{tab:rel_performance}B, except for GPT-4 (ICL), all approaches achieved higher overall F1 than human IAA. However, when longer-distance, inter-sentence relations are included (Table \ref{tab:rel_performance}A), performance is similar to IAA. This underscores the challenges of long-distance RE, where co-reference and varied relation expressions are involved. Such relations are also often indicated by section headers without explicit descriptions.

\subsection*{RE Error Analysis}
\begin{table}[]
\centering
\resizebox{\columnwidth}{!}{%
\begin{tabular}{llcccc|c}
\hline
\multirow{2}{*}{\textbf{\begin{tabular}[c]{@{}l@{}}Error\\ Types\end{tabular}}} & \multirow{2}{*}{\textbf{\begin{tabular}[c]{@{}l@{}}Error\\ Subtypes\end{tabular}}} & \multicolumn{4}{c|}{\textbf{\begin{tabular}[c]{@{}c@{}}All Relations\\ (Predicted Events)\end{tabular}}} & \textbf{\begin{tabular}[c]{@{}c@{}}Intra-sent Relations\\ (Gold Events)\end{tabular}} \\ \cline{3-7} 
 &  & \textbf{SpERT} & \textbf{PL-Marker} & \textbf{Flan-T5} & \textbf{llama 3 8B} & \textbf{GPT-4 (ICL)} \\ \hline
\multirow{5}{*}{\textbf{\begin{tabular}[c]{@{}l@{}}False\\ Negatives\end{tabular}}} & FN events & 6 & 4 & 23 & 11 & - \\
 & Relations implicated by symbols & 6 & 2 & 0 & 0 & 0 \\
 & Dense events & 10 & 5 & 4 & 9 & 8 \\
 & Other Intra-sentence relations & 2 & 14 & 2 & 0 & 1 \\
 & Other Inter-sentence relations & 15 & 12 & 9 & 5 & - \\ \cline{1-2}
\multirow{4}{*}{\textbf{\begin{tabular}[c]{@{}l@{}}False\\ Positives\end{tabular}}} & FP events & 4 & 2 & 8 & 3 & - \\
 & Dense events & 8 & 10 & 9 & 5 & 8 \\
 & Other Intra-sentence relations & 1 & 1 & 9 & 3 & 7 \\
 & Other Inter-sentence relations & 0 & 0 & 15 & 11 & - \\ \cline{1-2}
\textbf{\begin{tabular}[c]{@{}l@{}}Mis-classi-\\ fication\end{tabular}} & - & 4 & 7 & 8 & 5 & 7 \\ \cline{1-2}
\textbf{Overall} & - & 56 & 57 & 87 & 52 & 31 \\ \hline
\end{tabular}
}
\caption{Types of RE errors in 5 sampled notes from the test set. The 5 notes have 128 relations in total, and 87 of the 128 relations are intra-sentence. All GLMs' predictions are generated under the QA prompting format.}
\label{tab:re_error}
\end{table}

We conducted an error analysis for RE using the same 5 notes as in Table \ref{tab:ee_error} by manually characterizing the RE errors. We analyzed all RE errors for BERT-based LMs and GLMs with the QA format. For GPT-4 (ICL), we only analyzed errors associated with intra-sentence relations, as GPT-4 experimentation did not include inter-sentence relations. Table \ref{tab:re_error} presents a breakdown of RE error types. 
Cascading errors from EE, including FN and FP event trigger predictions, result in RE errors. 
Another major error source is from dense events with ambiguities and possibly hierarchical interactions. 
Consider the following example, `Significant \textit{hypotension} while hospitalized, consistent with \textit{sepsis}. Persisted despite \textit{antibiotics}.' PL-Marker captured the \textit{Worsen} relation between `\textit{antibiotics}' and `\textit{sepsis}', but missed the \textit{Worsen} relation between `\textit{antibiotics}' and `\textit{hypotension}' and the \textit{PIP} relation between `\textit{sepsis}' and `\textit{hypotension}.'

While BERT-based models generated more FNs, GLMs generated more FPs. The FNs for BERT-based models usually come from (1) relations implicated by punctuations, such as the sentence with a colon, `\textit{AKI}: Cr peaked to ..., likely due to supratherapeutic \textit{cyclosporine}', resulting in an FN associated with `\textit{cyclosporine}' \textit{Causes} `\textit{AKI}'; (2) inter-sentence relations where the multiple sentences discuss the same conditions, such as the sentences about the metastasis, `pelvis revealed extensive bony \textit{metastasis} ... \textit{Lupron} was given', resulting in the FN relation,  `\textit{Lupron}' \textit{AdminFor} `\textit{metastasis}'. FPs for GLMs are usually from (1) over-predicting relations when multiple events occur in close proximity, like a list of events; and (2) inter-sentence relations with the same trigger names, for example, `...  mild rash and groin \textit{pain} developing; monitor ... prescribed \textit{Oxycodone} prn for \textit{pain}', where the model must disambiguate which reference to `pain' is associated with the `Oxycodone' prescription.

The mis-classifications stem from \textit{Drug}-\textit{Problem} relations, where models often confuse \textit{Causes} with \textit{NotAdminBecause}. While the former indicates that the \textit{Problem} event is an adverse effect of the \textit{Drug}, the latter does not. Sentences with ambiguities, such as '\textit{Drug} \textit{A} was discontinued because of \textit{Problem} \textit{B}', require medical knowledge to distinguish between these two relations.



\section*{Discussion}

\subsection*{Drug Annotation Schema} \label{sec:drug_annotation_schema}
Previous studies on fine-grained drug annotations in clinical narratives, such as those reported in i2b2 2009 \citep{uzuner2010extracting}, n2c2 2018 \citep{henry20202018}, and MADE 1.0 \citep{jagannatha2019overview}, have explored the detailed characterization of \textit{Drug} events through attributes including dosage, route, and frequency. However, these attributes are not directly transferable to cancer treatment protocols, where drugs are typically administered as part of a regimen. For example, the R-CHOP regimen — a commonly used treatment for non-Hodgkin lymphoma (NHL) — combines multiple drugs (Rituximab, Cyclophosphamide, Doxorubicin Hydrochloride, Vincristine Sulfate, and Prednisone) each with specified dosages, routes, and frequencies that are synergistically designed to maximize therapeutic efficacy. Such regimens reflect a complex matrix of attributes that differ significantly from the singular drug attributes documented in traditional drug annotation tasks, thereby rendering these methods less applicable to oncology. Our approach of annotating \textit{Drug} event triggers as specific regimens can potentially be integrated with a hierarchical pharmacological knowledge base  \cite{malty2018computerized}. These knowledge bases are designed to characterize component medications and therapeutic contexts, among other elements, providing a structured framework for understanding and organizing complex oncology \cite{malty2018computerized}.

\subsection*{Comparison Across Models}
In EE and RE, comparing BERT models (SpERT and PL-Marker) and the top-performing GLM (Llama3) reveals no significant differences in overall end-to-end performance. In RE with gold standard triggers, Llama3-QA achieves the numerical highest performance. EE involves identifying specific textual spans and their relationships, while relation prediction focuses on reasoning about pre-identified events, relying on comprehensive language understanding. Although Llama3 performs similar to BERT models in trigger identification, it excels in understanding relationships between triggers. However, Llama3-QA requires a separate query for each potential relation, unlike BERT models that extract all relationships in a sentence in one step. This higher performance of Llama3 comes at an increased computational cost.

It is important to consider the extraction task requirements to balance performance benefits and computational costs. LLMs excel in tasks demanding medical knowledge and advanced reasoning and have achieved high performance in complex tasks, like medical exams \cite{singhal2023large}, common-sense reasoning \cite{labrak2024biomistral}, and dialogue summarization \cite{yim2023aci}. However, for tasks that require lower levels of abstraction, like identifying \textit{Drug} triggers, the enhanced capabilities of LLMs may be unnecessary, as smaller models like BERT and T5 can achieve comparable high performance. Future research could investigate the minimal scale of language models required to match human performance, especially for tasks that do not require high levels of abstraction. This would enable more computationally efficient models to achieve significant performance gains without the exponential increase in computational demands.


\subsection*{Considerations for Clinical Deployment}
Our experiments show that fine-grained \textit{Problem} and \textit{Drug} information can be extracted with similar performance to human annotators. Deploying such IE systems can enable large-scale,  real-time information usage in EHR-embedded clinical decision support applications and generate real-world evidence for learning health systems. This may lead to more effective, safe, and efficient care and a broader evidence base for decision-making at various levels in healthcare systems. 

Although the best-performing RE models achieve performance comparable to IAA, the overall IAA of 69.6 F1 indicates the inherent annotation challenges of this task.  To address false positives, it may be possible to incorporate confidence scores when annotating and predicting relations. BERT models can directly generate such scores as SoftMax probabilities, while GLMs can implement a second verification step to reduce false positives \cite{gero2023self}. False negatives are difficult to detect and may lead to severe consequences if adverse drug events are missed. An ensemble approach using multiple models may help capture more relations, but the effectiveness of this strategy depends on model quality and does not guarantee improved performance \cite{christopoulou2020adverse}. 

To facilitate large-scale deployment, the extracted \textit{Problem} and \textit{Drug} event triggers could be normalized to standardized medical lexicons like the International Classification of Diseases (ICD-10) \cite{world1992icd, ji2020bert, soroush2024large}. This normalization would enhance the integration of our IE systems into existing systems.

\section*{Conclusions}
This study presents the Clinical Concept Annotations for Cancer Events and Relations (CACER), a novel corpus that provides detailed annotations of medical problems, drugs, and their relationships from clinical narratives of oncology notes. Our baseline experiments with state-of-the-art transformer-based models achieved performance levels comparable to annotator agreement. Error analysis revealed several challenges facing current high-performing models: (1) enhancing the ability to generalize to unseen events for EE and (2) deciphering the complex context and medical knowledge required for RE. Future work will focus on integrating domain-specific knowledge into RE techniques and improving model capacity to accurately identify relationships across longer textual distances.

\section*{Acknowledgments} 
We express our sincere appreciation to all annotators for their contributions to the data set annotation: Nianiella Dorvall, Spencer Raub, and Emily Stiles. Furthermore, we sincerely thank Dr. Nic Dobbins from Johns Hopkins University for his invaluable insights and revisions that significantly enhanced this manuscript.

This is a pre-copy-editing, author-produced PDF of an article accepted for publication in JAMIA following peer review. The definitive publisher-authenticated version \cite{10.1093/jamia/ocae231} is available online at \url{https://academic.oup.com/jamia/advance-article/doi/10.1093/jamia/ocae231/7748302}.

\section*{Funding Statement}
This work was supported by the National Institutes of Health (NIH) - National Cancer Institute (Grant Nr. 1R01CA248422-01A1 and 1R21CA258242-01) and National Library of Medicine (Grant Nr. 2R15LM013209-02A1). The content is solely the responsibility of the authors and does not necessarily represent the official views of the NIH. 

\section*{Competing Interests Statement}
The authors have no competing interests to declare.

\section*{Contributorship Statement}
All authors contributed to dataset creation and IE method development. YF was responsible for dataset cleaning, conducting experiments, and analyzing errors. YF, GR, and KL wrote the initial draft, and all authors reviewed and revised the manuscript.

\section*{Data Availability Statement}
The code for baseline experiments and evaluation is available through our project GitHub: https://github.com/uw-bionlp/CACER. 

\bibliography{mybib}

\pagebreak

\newpage
\section*{Supplementary File}

\subsection*{Data Set Statistics} 
Table. \ref{tab:demographics} demonstrates the data set statistics and patient demographics.

\begin{table}[h]
\centering
\begin{tabular}{lcccc}
\hline
\textbf{Type} & \textbf{Subtype} & \textbf{Train} & \textbf{Valid} & \textbf{Test} \\ \hline
\textbf{\# Notes} & - & 400 & 60 & 115 \\ \cline{1-2}
\textbf{\# Unique patients} & - & 306 & 43 & 115 \\ \cline{1-2}
\multirow{2}{*}{\textbf{Annotation}} & Double & 0 & 47 & 115 \\
 & Single & 400 & 13 & 0 \\ \cline{1-2}
\multirow{2}{*}{\textbf{Cancer}} & DLBCL & 193 & 22 & 56 \\
 & Prostate & 207 & 38 & 61 \\ \cline{1-2}
\multirow{2}{*}{\textbf{Gender}} & Male & 314 & 51 & 92 \\
 & Female & 86 & 9 & 23 \\ \hline
\end{tabular}
\caption{Note statistics and patient demographics.}
\label{tab:demographics}
\end{table}

\subsection*{Prompts}
All task prompts are provided as `\textit{user}' messages in GLMs.

\noindent
\textbf{Event Extraction}

``You are a medical expert. Extract all drug and medical problem events from the following clinical note. All events constraints span-only arguments and/or valued arguments. Span-only arguments must use the span original from the clinical note. A medical problem event contains required arguments as a trigger span and an assertion value (present, absent, possible, conditional, hypothetical, not\_patient), as well as optional arguments as at most one anatomy span, at most one duration span, at most one frequency span, characteristics spans, change value (no\_change, improving, worsening, resolved), severity value (mild, moderate, severe). The drug event contains a required argument as a trigger span.''

\noindent
\textbf{Relation Extraction, Marker Format}

``Extract all relations related to drug and medical problems from this clinical note.

Clinical Note: `...'''

\noindent
\textbf{Relation Extraction, QA Format 1 for Drug-Problem relations}

``What is the relationship between \textless \textit{A} \textgreater (Drug) and \textless \textit{B} \textgreater (Problem) in the following clinical notes? 

Clinical Note: `...'

Options:

(A) \textit{A} is given as a treatment for \textit{B},  but the outcome is not mentioned in the sentence.

(B) \textit{A} is not given or discontinued because of the \textit{B} that it does not cause.

(C) \textit{A} is given as a treatment for \textit{B},  but \textit{A} does not cure the \textit{B}, does not improve the \textit{B}, or makes the \textit{B} worse.

(D) \textit{A} is not given as a treatment for \textit{B}, but it causes \textit{B}.

(E) \textit{A} improves, cures, stablize \textit{B}.\

(F) None of the above.
''

\noindent
\textbf{Relation Extraction, QA Format 2 for Problem-Problem relations}

``What is the relationship between \textless \textit{A} \textgreater (Problem) and \textless \textit{B} \textgreater (Problem) in the following clinical notes? 

Clinical Note: `...'

Options:

(A) \textit{A} causes, describes or reveals aspects of \textit{B}.

(B) \textit{B} causes, describes or reveals aspects of \textit{A}. 

(C) None of the above.''

\subsection*{Condensed Annotation Guideline} 
You are a medical expert. Extract all drug and medical problem events from the following clinical note. All events contains a trigger, span-only arguments and/or valued arguments. Trigger and span-only arguments must use the original span from the clinical note, and the shortest span possible. Valued arguments must be chosen from a pre-defined list.
For every note, output None, if the span or value does not exist.
Output the events by the order of trigger occurrence from clinical note. If there are multiple arguments of the same type, separate them by $\textless  s\textgreater$. For example, ‘congestive $\textless  s\textgreater$ progressive’
Multiple Drug and Problem events are separated by [SEP]

This is an example format:
\textless Problem\textgreater {span} \textless Assertion\textgreater {value} \textless Anatomy\textgreater {span} $\textless  s\textgreater$.. $\textless  s\textgreater$ {span} \textless Duration\textgreater  {span} \textless Frequency\textgreater  {span} \textless Characteristics\textgreater  {span} $\textless  s\textgreater$.. $\textless  s\textgreater$ {span}  \textless Change\textgreater  {value} \textless Severity\textgreater {value} [SEP] \textless Drug\textgreater  {span}  [SEP] … 

The drug event contains only required argument as a trigger span, which is the shortest span possible indicating a drug or treatment name. 

A medical problem event contains required arguments as a trigger span and an assertion value (present, absent, possible, conditional, hypothetical, not$\textunderscore$patient). The problem trigger span is be the shortest span possible. The problem trigger is a span that contains observations made by patients or clinicians about the patient’s body or mind that are thought to be abnormal or caused by a disease. Generally, the trigger span should not include anatomical information or characteristics of the problem, as this information is captured through separate Anatomy and Characteristics arguments. They are loosely based on the UMLS semantic types of pathologic functions, disease, or syndrome, mental or behavioral dysfunction, cell or molecular dysfunction, congenital abnormality, acquired abnormality, injury or poisoning, anatomic abnormality, neoplastic process, virus/bacterium, sign or symptom, but are not limited by UMLS coverage. 

\begin{enumerate}[topsep=0pt,itemsep=-1ex,partopsep=1ex,parsep=1ex]
    \item present: patient experienced or is experiencing 
    \item absent: patient has not or is not experiencing
    \item possible: patient may be experiencing (denoted by terms like “probably” or “likely”)
    \item conditional: patient only experiences under specific conditions
    \item hypothetical: patient may experience in the future
    \item not$\textunderscore$patient: not associated with the patient
\end{enumerate}
The problem event has the following optional arguments:
\begin{enumerate}
    \item Anatomy (span): indicates the body part or region of the body associated with the problem.
    \item Duration (span): how long a problem has persisted or when the problem started.
    \item Frequency (span): how often a problem occurs (e.g. occasionally, intermittently, chronic, daily, hourly, persistent, etc.)
    \item Characteristics (span): problem descriptors, including descriptions of color, consistency, sound, pain, diffuse/localized, etc. A single event (trigger) may have multiple Characteristics spans. For example, a cough could be described through two Characteristics spans, like “dry non-productive” and “painful.”
    \item Change (value): captures explicit descriptions of changes in the state of the problem. Choose from no$\textunderscore$change, improved, worsened, and resolved.
    \item Severity (value): Choose from mild, moderate, severe. Severity can be direct description of the patient status such as ‘mild fever’, or inferred by the treatment plan as (1). No treatment needed - mild, (2). treatment needed - moderate (3). hospitalization needed - severe
\end{enumerate}

\subsection*{Relation Extraction Context Window}

The RE model can be constrained by their context length, ranging from 512 tokens in BERT, 1024 tokens in Flan-T5, to 32k tokens for GPT-4. In this work, we consider a context window for a certain relation as the minimum continuous sentence span that contains both head and tail events. Our RE approaches are restricted by context windows no longer than 5 sentences and 400 Bio+Clinical BERT tokens, which contain the majority (98.7\%) of the relations. Intra-sentence relations constitute 70.7\% of the relations from this data set. 




GLMs classify each possible relation pair within its corresponding context window. On the other hand, BERT-based models process each unique context window only once. During the training phase, the context window encompasses all events and relations. During inference, to ensure that each relation is predicted only once, predictions are considered valid only when meeting one of the following criteria: (1) the head trigger appears in the beginning sentence and the tail trigger in the ending sentence, (2) the tail trigger appears in the beginning sentence and the head trigger in the ending sentence, or (3) the context window consists of a single sentence.

\end{document}